\documentclass{article} % For LaTeX2e
\usepackage{iclr2017_conference,times}
\usepackage{hyperref,url}
\usepackage{amsmath,amssymb}
\usepackage{graphicx,soul,subcaption}
\usepackage{chngcntr}
\usepackage{siunitx}

\title{Deep Unsupervised Clustering with Gaussian Mixture Variational Autoencoders}

\author{Nat Dilokthanakul$^{1,*}$, Pedro A. M. Mediano$^1$, Marta Garnelo$^1$,\\
\textbf{Matthew C. H. Lee$^1$, Hugh Salimbeni$^1$, Kai Arulkumaran$^2$ \& Murray Shanahan$^1$}\\
$^1$Department of Computing, $^2$Department of Bioengineering\\
Imperial College London\\
London, UK\\
$^*$\texttt{n.dilokthanakul14@imperial.ac.uk}
}

\begin{document}

\maketitle

\begin{abstract}

We study a variant of the variational autoencoder model (VAE) with a Gaussian mixture as a prior distribution, with the goal of performing unsupervised clustering through deep generative models. We observe that the known problem of over-regularisation that has been shown to arise in regular VAEs also manifests itself in our model and leads to cluster degeneracy. We show that a heuristic called minimum information constraint that has been shown to mitigate this effect in VAEs can also be applied to improve unsupervised clustering performance with our model. Furthermore we analyse the effect of this heuristic and provide an intuition of the various processes with the help of visualizations. Finally, we demonstrate the performance of our model on synthetic data, MNIST and SVHN, showing that the obtained clusters are distinct, interpretable and result in achieving competitive performance on unsupervised clustering to the state-of-the-art results. 

\end{abstract}

%%%%%%%%%%%%%%%%%%%%%%%%%%%%%%%%%%%%%%%%%%%%%%%%
%%%%%%%%%%%%%%%%%%%%%%%%%%%%%%%%%%%%%%%%%%%%%%%%
\section{Introduction}

Unsupervised clustering remains a fundamental challenge in machine learning research. While long-established methods such as $k$-means and Gaussian mixture models (GMMs) \citep{bishop2006pattern} still lie at the core of numerous applications \citep{aggarwal2013data}, their similarity measures are limited to local relations in the data space and are thus unable to capture hidden, hierarchical dependencies in latent spaces. Alternatively, deep generative models can encode rich latent structures. While they are not often applied \emph{directly} to unsupervised clustering problems, they can be used for dimensionality reduction, with classical clustering techniques applied to the resulting low-dimensional space \citep{xie2015unsupervised}. This is an unsatisfactory approach as the assumptions underlying the dimensionality reduction techniques are generally independent of the assumptions of the clustering techniques.

Deep generative models try to estimate the density of observed data under some assumptions about its latent structure, i.e., its hidden causes. They allow us to reason about data in  more complex ways than in models trained purely through supervised learning. However, inference in models with complicated latent structures can be difficult. Recent breakthroughs in approximate inference have provided  tools for constructing tractable inference algorithms. As a result of combining differentiable models with variational inference, it is possible to scale up inference to datasets of sizes that would not have been possible with earlier inference methods \citep{rezende2014stochastic}. One popular algorithm under this framework is the variational autoencoder (VAE) \citep{kingma2013auto, rezende2014stochastic}.

In this paper, we propose an algorithm to perform unsupervised clustering within the VAE framework. To do so, we postulate that generative models can be tuned for unsupervised clustering by making the assumption that the observed data is generated from a multimodal prior distribution, and, correspondingly, construct an inference model that can be directly optimised using the reparameterization trick. We also show that the problem of over-regularisation in VAEs can severely effect the performance of clustering, and that it can be mitigated with the minimum information constraint introduced by \cite{kingma2016improving}.

%%%%%%%%%%%%%%%%%%%%%%%%%%%%%%%%%%%%%%%%%%%%%%%%
\subsection{Related Work}

Unsupervised clustering can be considered a subset of the problem of disentangling latent variables, which aims to find structure in the latent space in an unsupervised manner. Recent efforts have moved towards training models with disentangled latent variables corresponding to different factors of variation in the data. Inspired by the learning pressure in the ventral visual stream, \cite{higgins2016early} were able to extract disentangled features from images by adding a regularisation coefficient to the lower bound of the VAE. As with VAEs, there is also effort going into obtaining disentangled features from generative adversarial networks (GANs) \citep{goodfellow2014generative}. This has been recently achieved with InfoGANs \citep{chen2016infogan}, where structured latent variables are included as part of the noise vector, and the mutual information between these latent variables and the generator distribution is then maximised as a mini-max game between the two networks. Similarly, Tagger \citep{greff2016tagger}, which combines iterative amortized grouping and ladder networks, aims to perceptually group objects in images by iteratively denoising its inputs and assigning parts of the reconstruction to different groups. \cite{johnson2016composing} introduced a way to combine amortized inference with stochastic variational inference in an algorithm called structured VAEs. Structured VAEs are capable of training deep models with GMM as prior distribution. \cite{shu2016stochastic} introduced a VAE with a multimodal prior where they optimize the variational approximation to the standard variational objective showing its performance in video prediction task.

The work that is most closely related to ours is the stacked generative semi-supervised model (M1+M2) by \cite{kingma2014semi}. One of the main differences is the fact that their prior distribution is a neural network transformation of both continuous and discrete variables, with Gaussian and categorical priors respectively. The prior for our model, on the other hand, is a neural network transformation of Gaussian variables, which parametrise the means and variances of a mixture of Gaussians, with categorical variables for the mixture components. Crucially, \cite{kingma2014semi} apply their model to semi-supervised classification tasks, whereas we focus on unsupervised clustering. Therefore, our inference algorithm is more specific to the latter.

We compare our results against several orthogonal state-of-the-art techniques in unsupervised clustering with deep generative models: deep embedded clustering (DEC) \citep{xie2015unsupervised}, adversarial autoencoders (AAEs) \citep{makhzani2015adversarial} and categorial GANs (CatGANs) \citep{springenberg2015unsupervised}.

%%%%%%%%%%%%%%%%%%%%%%%%%%%%%%%%%%%%%%%%%%%%%%%%
%%%%%%%%%%%%%%%%%%%%%%%%%%%%%%%%%%%%%%%%%%%%%%%%
\section{Variational Autoencoders}
\label{vae}

VAEs are the result of combining variational Bayesian methods with the flexibility and scalability provided by neural networks \citep{kingma2013auto, rezende2014stochastic}. Using variational inference it is possible to turn intractable inference problems into optimisation problems \citep{wainwright2008graphical}, and thus expand the set of available tools for inference to include optimisation techniques as well. Despite this, a key limitation of classical variational inference is the need for the likelihood and the prior to be conjugate in order for most problems to be tractably optimised, which in turn can limit the applicability of such algorithms. Variational autoencoders introduce the use of neural networks to output the conditional posterior \citep{kingma2013auto} and thus allow the variational inference objective to be tractably optimised via stochastic gradient descent and standard backpropagation. This technique, known as the reparametrisation trick, was proposed to enable backpropagation through continuous stochastic variables. While under normal circumstances backpropagation through stochastic variables would not be possible without Monte Carlo methods, this is bypassed by constructing the latent variables through the combination of a deterministic function and a separate source of noise. We refer the reader to \cite{kingma2013auto} for more details.

%%%%%%%%%%%%%%%%%%%%%%%%%%%%%%%%%%%%%%%%%%%%%%%%
%%%%%%%%%%%%%%%%%%%%%%%%%%%%%%%%%%%%%%%%%%%%%%%%
\section{Gaussian Mixture Variational Autoencoders}
\label{sec:gmvae}

In regular VAEs, the prior over the latent variables is commonly an isotropic Gaussian. This choice of prior causes each dimension of the multivariate Gaussian to be pushed towards learning a separate continuous factor of variation from the data, which can result in learned representations that are structured and disentangled. While this allows for more interpretable latent variables \citep{higgins2016early}, the Gaussian prior is limited because the learnt representation can only be unimodal and does not allow for more complex representations. As a result, numerous extensions to the VAE have been developed, where more complicated latent representations can be learned by specifying increasingly complex priors~\citep{chung2015arecurrent, gregor2015draw, eslami2016attend}.

In this paper we choose a mixture of Gaussians as our prior, as it is an intuitive extension of the unimodal Gaussian prior. If we assume that the observed data is generated from a mixture of Gaussians, inferring the class of a data point is equivalent to inferring which mode of the latent distribution the data point was generated from. While this gives us the possibility to segregate our latent space into distinct classes, inference in this model is non-trivial. It is well known that the reparametrisation trick which is generally used for VAEs cannot be directly applied to discrete variables. Several possibilities for estimating the gradient of discrete variables have been proposed \citep{glynn1990likelihood, titsias2015local}. \cite{graves2016stochastic} also suggested an algorithm for backpropagation through GMMs. Instead, we show that by adjusting the architecture of the standard VAE, our estimator of the variational lower bound of our Gaussian mixture variational autoencoder (GMVAE) can be optimised with standard backpropagation through the reparametrisation trick, thus keeping the inference model simple. 

%%%%%%%%%%%%%%%%%%%%%%%%%%%%%%%%%%%%%%%%%%%%%%%%
\subsection{Generative and Recognition Models}

Consider the generative model $p_{\beta, \theta}(\pmb{y},\pmb{x},\pmb{w},\pmb{z}) = p(\pmb{w})p(\pmb{z})p_{\beta}(\pmb{x}|\pmb{w},\pmb{z})p_{\theta}(\pmb{y}|\pmb{x})$, where an observed sample $\pmb{y}$ is generated from a set of latent variables $\pmb{x}$, $\pmb{w}$ and $\pmb{z}$ under the following process:
\begin{subequations}
\begin{align}
  \pmb{w} &\sim \mathcal{N}(0, \pmb{I})  \\
  \pmb{z} &\sim Mult(\pmb{\pi}) \\
  \pmb{x}|\pmb{z},\pmb{w} &\sim \prod_{k=1}^K \mathcal{N}\left(\pmb{\mu}_{z_{k}}(\pmb{w}; \beta), diag\left(\pmb{\sigma}^2_{z_{k}}(\pmb{w}; \beta)\right) \right)^{z_k} \\
  \pmb{y}|\pmb{x} &\sim \mathcal{N}\left(\pmb{\mu}(\pmb{x}; \theta), diag\left(\pmb{\sigma}^2(\pmb{x}; \theta)\right)\right) \text{ or } \mathcal{B}(\pmb{\mu}(\pmb{x}; \theta)) ~ .
\end{align}
\end{subequations}
\noindent where $K$ is a predefined number of components in the mixture, and $\pmb{\mu}_{z_k}(\cdot; \beta), \pmb{\sigma}^2_{z_k}(\cdot; \beta), \pmb{\mu}(\cdot; \theta)$, and $\pmb{\sigma}^2(\cdot; \theta)$ are given by neural networks with parameters $\beta$ and $\theta$, respectively. That is, the observed sample $\pmb{y}$ is generated from a neural network observation model parametrised by $\theta$ and the continuous latent variable $\pmb{x}$. Furthermore, the distribution of $\pmb{x}|\pmb{w}$ is a Gaussian mixture with means and variances specified by another neural network model parametrised by $\beta$ and with input $\pmb{w}$.

\begin{figure}[h]
\centering
\begin{subfigure}{0.5\textwidth}
	\centering
	\includegraphics[trim={8.4cm 12cm 8.4cm 12.2cm},clip,width = 0.8\textwidth]{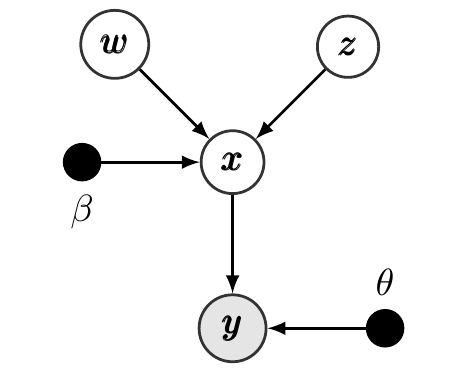}
\end{subfigure}%
\begin{subfigure}{0.5\textwidth}
	\centering
	\includegraphics[trim={8.4cm 12cm 8.4cm 12.2cm},clip,width = 0.8\textwidth]{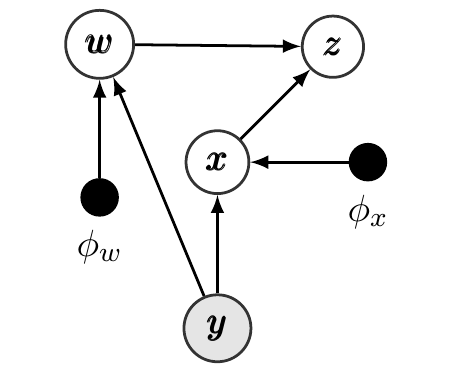}
\end{subfigure}%
\caption{Graphical models for the Gaussian mixture variational autoencoder (GMVAE) showing the generative model (left) and the variational family (right).}
\label{fig:graphModel}
\end{figure}

More specifically, the neural network parameterised by $\beta$ outputs a set of $K$ means $\pmb{\mu}_{z_k}$ and $K$ variances $\pmb{\sigma}^2_{z_k}$, given $\pmb{w}$ as input. A one-hot vector $\pmb{z}$ is sampled from the mixing probability $\pmb{\pi}$, which chooses one component from the Gaussian mixture. We set the parameter $\pi_k = K^{-1}$ to make $\pmb{z}$ uniformly distributed. The generative and variational views of this model are depicted in Fig.~\ref{fig:graphModel}.

%%%%%%%%%%%%%%%%%%%%%%%%%%%%%%%%%%%%%%%%%%%%%%%%
\subsection{Inference with the Recognition Model}

The generative model is trained with the variational inference objective, i.e. the log-evidence lower bound (ELBO), which can be written as
\begin{align}
 \mathcal{L}_{ELBO} = \mathbb{E}_q\bigg[ \frac{ p_{\beta, \theta}(\pmb{y},\pmb{x},\pmb{w},\pmb{z}) }{q(\pmb{x},\pmb{w},\pmb{z} | \pmb{y} )}  \bigg].
\end{align}
We assume the mean-field variational family $q(\pmb{x},\pmb{w},\pmb{z}| \pmb{y})$ as a proxy to the posterior which factorises as $q(\pmb{x},\pmb{w},\pmb{z}| \pmb{y}) = \prod_i q_{\phi_{x}}(\pmb{x}_i| \pmb{y}_i)q_{\phi_w}(\pmb{w}_i| \pmb{y}_i)p_{\beta}(\pmb{z}_i| \pmb{x}_i, \pmb{w}_i)$, where $i$ indexes over data points. To simplify further notation, we will drop $i$ and consider one data point at a time. We parametrise each variational factor with the recognition networks $\phi_{x}$ and $\phi_{w}$ that output the parameters of the variational distributions and specify their form to be Gaussian posteriors. We derived the $z$-posterior, $p_{\beta}(\pmb{z}| \pmb{x}, \pmb{w})$, as:
\begin{align}
 p_{\beta}(z_j= 1|\pmb{x}, \pmb{w}) &= \frac{p(z_j = 1)p(\pmb{x}|z_j = 1, \pmb{w})}{\sum_{k=1}^K p(z_k =1)p(\pmb{x}|z_j = 1, \pmb{w})} \nonumber \\
                   &=  \frac{\pi_j \mathcal{N}(\pmb{x}|\mu_j(\pmb{w};\beta), \sigma_j(\pmb{w};\beta))}{\sum_{k=1}^K \pi_k \mathcal{N}(\pmb{x}|\mu_k(\pmb{w};\beta), \sigma_k(\pmb{w};\beta))} ~ .
\label{eq:zposterior}
\end{align}

The lower bound can then be written as,
\begin{equation}
\begin{aligned}
 \mathcal{L}_{ELBO} &= \mathbb{E}_{q(\pmb{x}|\pmb{y})}\big[ \log{p}_{\theta}(\pmb{y}|\pmb{x}) \big] - \mathbb{E}_{q(\pmb{w}| \pmb{y})p(\pmb{z}| \pmb{x}, \pmb{w})}\big[ KL(q_{\phi_x}(\pmb{x}| \pmb{y})|| p_{\beta}(\pmb{x}|\pmb{w},\pmb{z})) \big] \\
 			 &\quad - KL(q_{\phi_w}(\pmb{w}| \pmb{y})|| p(\pmb{w})) -  \mathbb{E}_{q(\pmb{x}|\pmb{y})q(\pmb{w}|\pmb{y})}\big[ KL(p_{\beta}(\pmb{z}| \pmb{x},\pmb{w})|| p(\pmb{z})) \big].
  \label{eq:elbo}%
\end{aligned}
\end{equation}
We refer to the terms in the lower bound as the reconstruction term, conditional prior term, $w$-prior term and $z$-prior term respectively.

%%%%%%%%%%%%%%%%%%%%%%%%%%%%%%%%%%%%%%%%%%%%%%%%
%%%%%%%%%%%%%%%%%%%%%%%%%%%%%%%%%%%%%%%%%%%%%%%%
\subsubsection{The Conditional Prior Term}
The reconstruction term can be estimated by drawing Monte Carlo samples from $q(\pmb{x}|\pmb{y})$, where the gradient can be backpropagated with the standard reparameterisation trick ~\citep{kingma2013auto}. The $w$-prior term can be calculated analytically.

Importantly, by constructing the model this way, the conditional prior term can be estimated using Eqn.~\ref{eq:conditionalPrior} without the need to sample from the discrete distribution $p(\pmb{z}|\pmb{x}, \pmb{w})$.
\begin{equation}
\begin{aligned}
 	\mathbb{E}_{q(\pmb{w}| \pmb{y})p(\pmb{z}| \pmb{x}, \pmb{w})}\Big[ KL\big(q_{\phi_x}(\pmb{x}| \pmb{y})|| p_{\beta}(\pmb{x}|\pmb{w},\pmb{z})\big)\Big] \approx \qquad \qquad\\
 	\frac{1}{M}\sum_{j=1}^M \sum_{k=1}^K p_{\beta}(z_k = 1 | \pmb{x}^{(j)},\pmb{w}^{(j)}) KL\left(q_{\phi_x}(\pmb{x}| \pmb{y})|| p_{\beta}(\pmb{x}| \pmb{w}^{(j)}, z_k = 1)\right)
\end{aligned}
\label{eq:conditionalPrior}
\end{equation}

Since $p_{\beta}(\pmb{z}| \pmb{x}, \pmb{w})$ can be computed for all $\pmb{z}$ with one forward pass, the expectation over it can be calculated in a straightforward manner and backpropagated as usual. The expectation over $q_{\phi_w}(\pmb{w}| \pmb{y})$ can be estimated with $M$ Monte Carlo samples and the gradients can be backpropagated via the reparameterisation trick. This method of calculating the expectation is similar to the marginalisation approach of ~\cite{kingma2014semi}, with a subtle difference. ~\cite{kingma2014semi} need multiple forward passes to obtain each component of the $z$-posterior. Our method requires wider output layers of the neural network parameterised by $\beta$, but only need one forward pass. Both methods scale up linearly with the number of clusters.

%%%%%%%%%%%%%%%%%%%%%%%%%%%%%%%%%%%%%%%%%%%%%%%%
%%%%%%%%%%%%%%%%%%%%%%%%%%%%%%%%%%%%%%%%%%%%%%%%

\subsection{The KL Cost of the Discrete Latent Variable}

The most unusual term in our ELBO is the $z$-prior term. The $z$-posterior calculates the clustering assignment probability directly from the value of $x$ and $w$, by asking how far $x$ is from each of the cluster positions generated by $w$. Therefore, the $z$-prior term can reduce the KL divergence between the $z$-posterior and the uniform prior by concurrently manipulating the position of the clusters and the encoded point $x$. Intuitively, it would try to merge the clusters by maximising the overlap between them, and moving the means closer together. This term, similar to other KL-regularisation terms, is in tension with the reconstruction term, and is expected to be over-powered as the amount of training data increases.

\subsection{The Over-regularisation Problem}

The possible overpowering effect of the regularisation term on VAE training has been described numerous times in the VAE literature \citep{bowman2015generating,sonderby2016train,kingma2016improving,chen2016variational}. As a result of the strong influence of the prior, the obtained latent representations are often overly simplified and poorly represent the underlying structure of the data.
So far there have been two main approaches to overcome this effect: one solution is to anneal the KL term during training by allowing the reconstruction term to train the autoencoder network before slowly incorporating the regularization from the KL term \citep{sonderby2016train}. The other main approach involves modifying the objective function by setting a cut-off value that removes the effect of the KL term when it is below a certain threshold \citep{kingma2016improving}.
As we show in the experimental section below, this problem of over-regularisation is also prevalent in the assignment of the GMVAE clusters and manifests itself in large degenerate clusters. While we show that the second approach suggested by \cite{kingma2016improving} does indeed alleviate this merging phenomenon, finding solutions to the over-regularization problem remains a challenging open problem.

%However, VAE are known to have difficulty during optimisation where the regularisation term influence the optimisation too strongly resulting in convergence on poor minimum result in the latent representation are over-simplified. This problem are observed many times (cite, cite, cite) with the KL cost term `turn-off' many of the latent variables where .... . There are two main ideas to help with the problem. (ladder cite) propose annealing of the KL term by letting the reconstruction term strengthen the encoderer/decoder network before slowly applying the KL regularisation. (Kingma cite) propose a modification to the objective by stopping the KL to regularise too much. In experimental section, we will show that the problem of over-regularisation also happen in the cluster assignment of the GMVAE, however, in our case the problem is the clusters are being merged too strongly. we will show that the (kingma)'s free-bit hack helps with our problem of degenerate clusters. Solving over-regularise problem is still an interesting open problem.

\section{Experiments}

The main objective of our experiments is not only to evaluate the accuracy of our proposed model, but also to understand the optimisation dynamics involved in the construction of meaningful, differentiated latent representations of the data. This section is divided in three parts:

\begin{enumerate}
\item We first study the inference process in a low-dimensional synthetic dataset, and focus in particular on how the over-regularisation problem affects the clustering performance of the GMVAE and how to alleviate the problem;

\item We then evaluate our model on an MNIST unsupervised clustering task; and

\item We finally show generated images from our model, conditioned on different values of the latent variables, which illustrate that the GMVAE can learn disentangled, interpretable latent representations.
\end{enumerate}

Throughout this section we make use of the following datasets:

\begin{itemize}
\item \textbf{Synthetic data}: We create a synthetic dataset mimicking the presentation of \cite{johnson2016composing}, which is a 2D dataset with 10,000 data points created from the arcs of 5 circles.

\item \textbf{MNIST}: The standard handwritten digits dataset, composed of 28x28 grayscale images and consisting of 60,000 training samples and 10,000 testing samples \citep{lecun1998gradient}.

\item \textbf{SVHN}: A collection of 32x32 images of house numbers \citep{netzer2011reading}. We use the cropped version of the standard and the extra training sets, adding up to a total of approximately 600,000 images.
\end{itemize}

\subsection{Synthetic data}

We quantify clustering performance by plotting the magnitude of the $z$-prior term described in Eqn.~\ref{eq:z-prior} during training. This quantity can be thought of as a measure of how much different clusters overlap. Since our goal is to achieve meaningful clustering in the latent space, we would expect this quantity to go down as the model learns the separate clusters.

\begin{equation}
\begin{aligned}
 \mathcal{L}_{z} &= -  \mathbb{E}_{q(\pmb{x}|\pmb{y})q(\pmb{w}|\pmb{y})}\big[ KL(p_{\beta}(\pmb{z}| \pmb{x},\pmb{w})|| p(\pmb{z})) \big]
  \label{eq:z-prior}%
\end{aligned}
\end{equation}

\begin{figure}[h]
\centering
	\begin{subfigure}{.33\textwidth}
	  \centering
	  \includegraphics[width=1.0\linewidth]{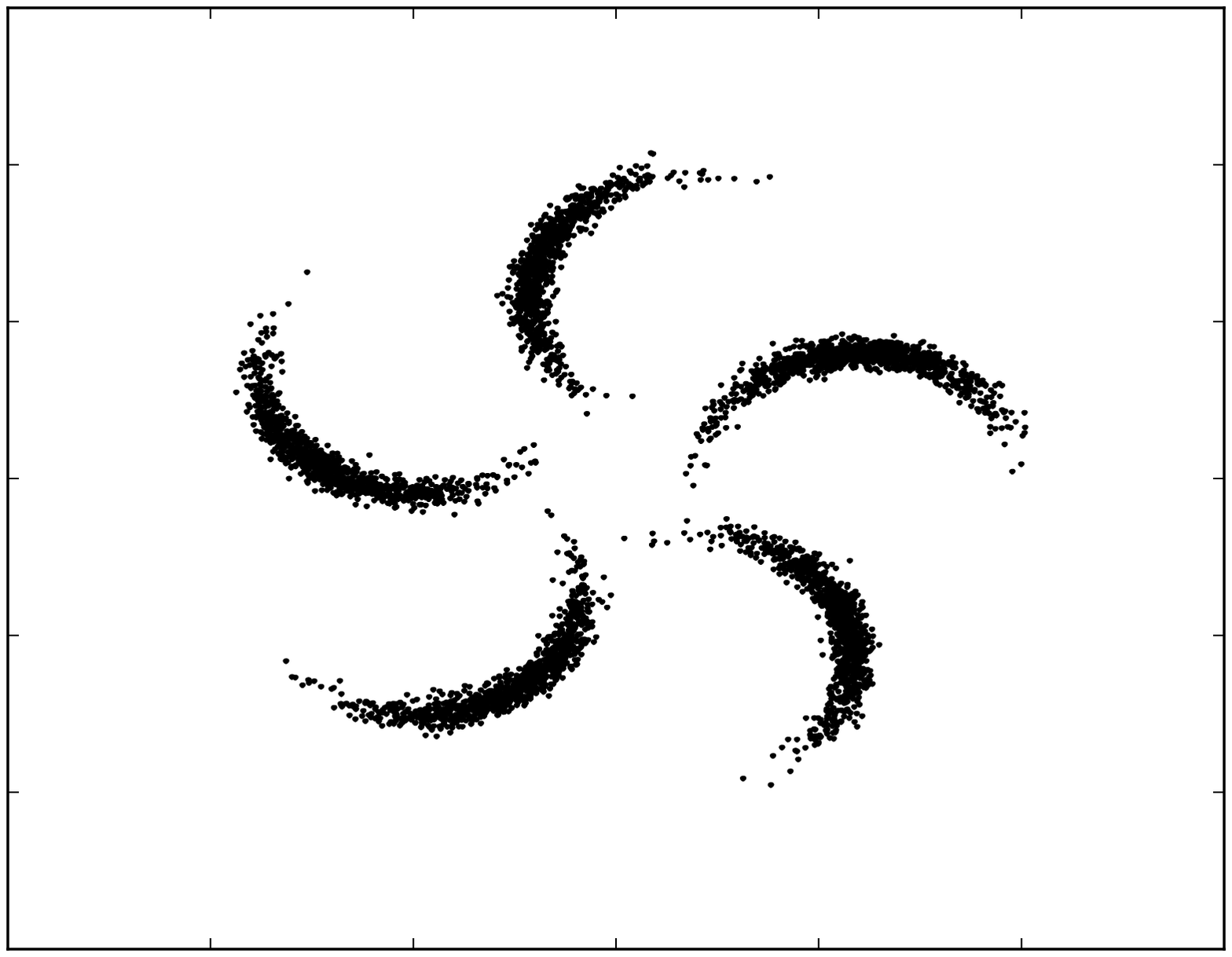}
	  \caption{Data points in data space}
      \label{fig:toyData}
	\end{subfigure}%
	\begin{subfigure}{.33\textwidth}
	  \centering
	  \includegraphics[width=1.0\linewidth]{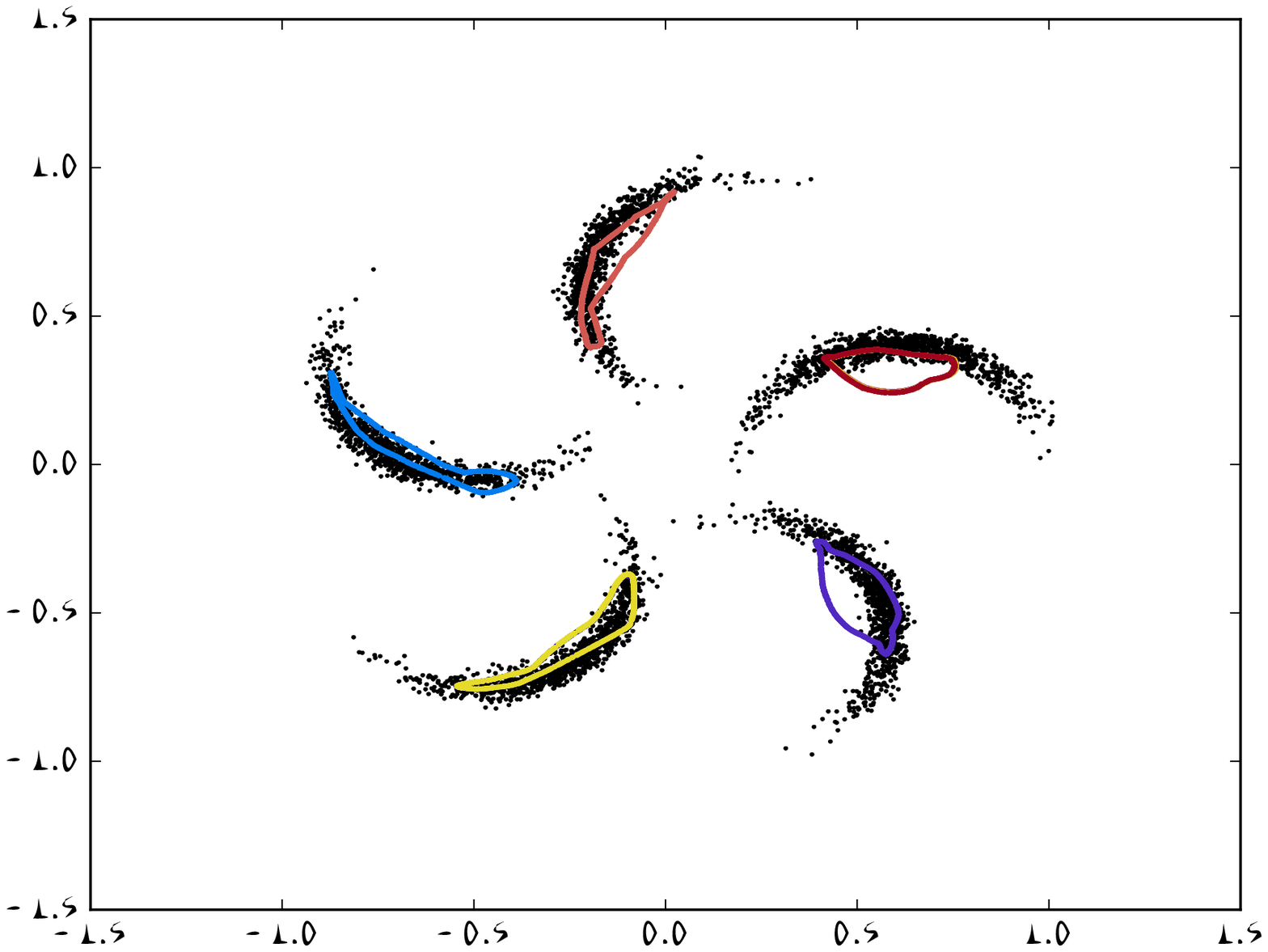}
	  \caption{Density of GMVAE}
      \label{fig:gmvaeDensity}
	\end{subfigure}%
	\begin{subfigure}{.33\textwidth}
	  \centering
	  \includegraphics[width=1.0\linewidth]{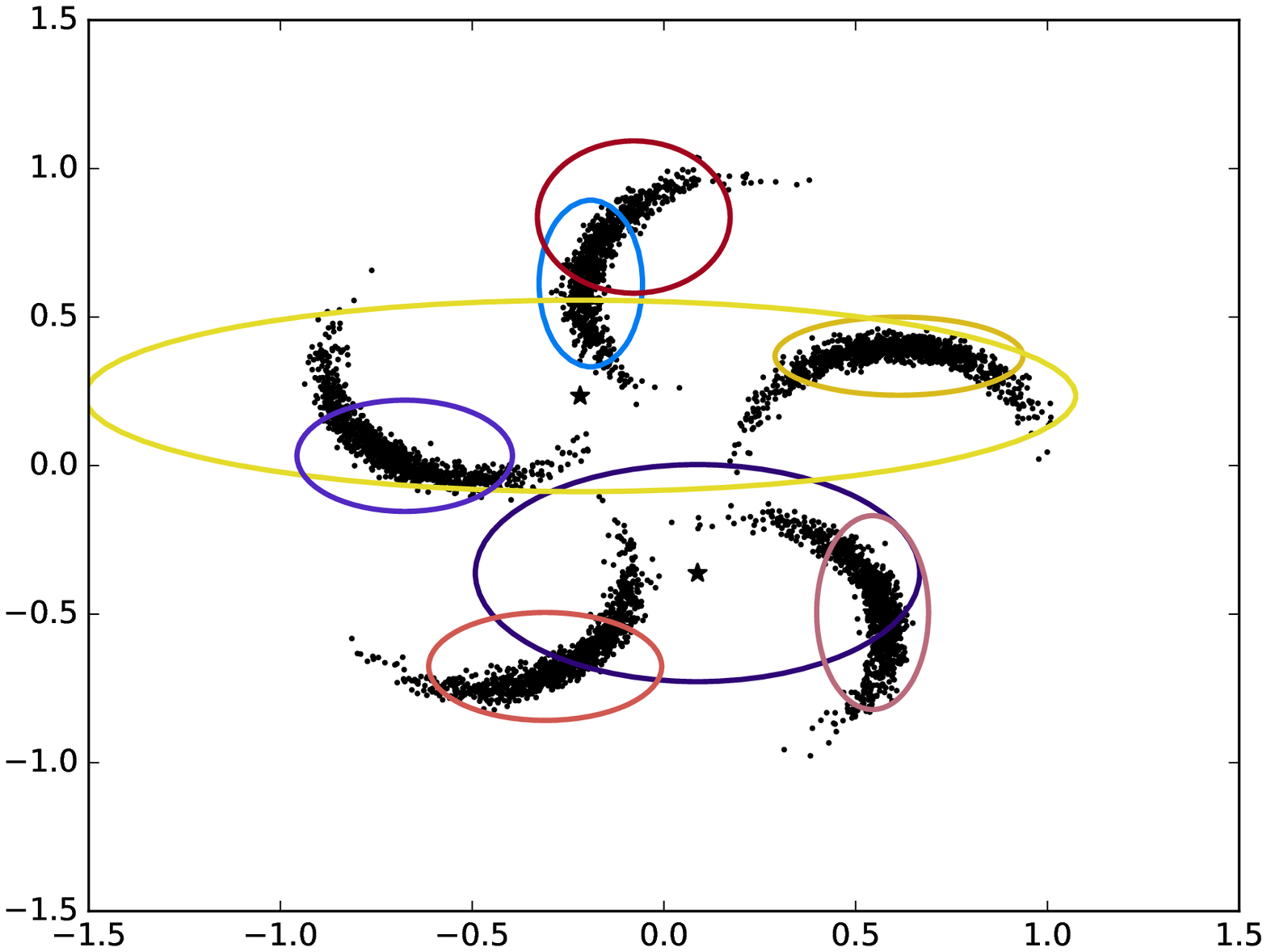}
	  \caption{Density of GMM}
	  \label{fig:gmmDensity}
	\end{subfigure}

	\begin{subfigure}{.33\textwidth}
    \centering
	  \includegraphics[width=1.0\linewidth]{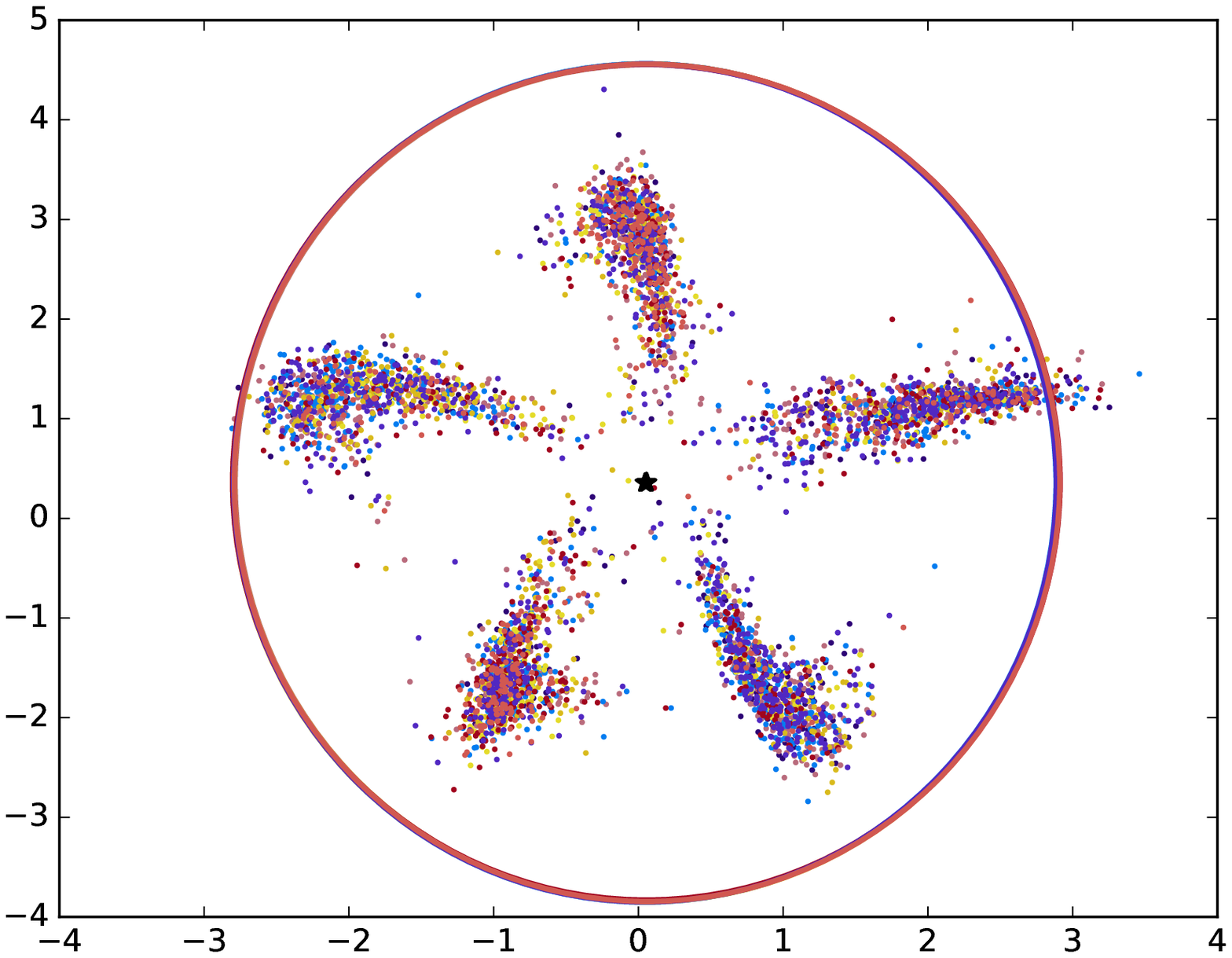}
	  \caption{Latent space, at poor optimum}
	  \label{fig:poorOptimal}
	\end{subfigure}%
	\begin{subfigure}{.33\textwidth}
    \centering
	  \includegraphics[width=1.0\linewidth]{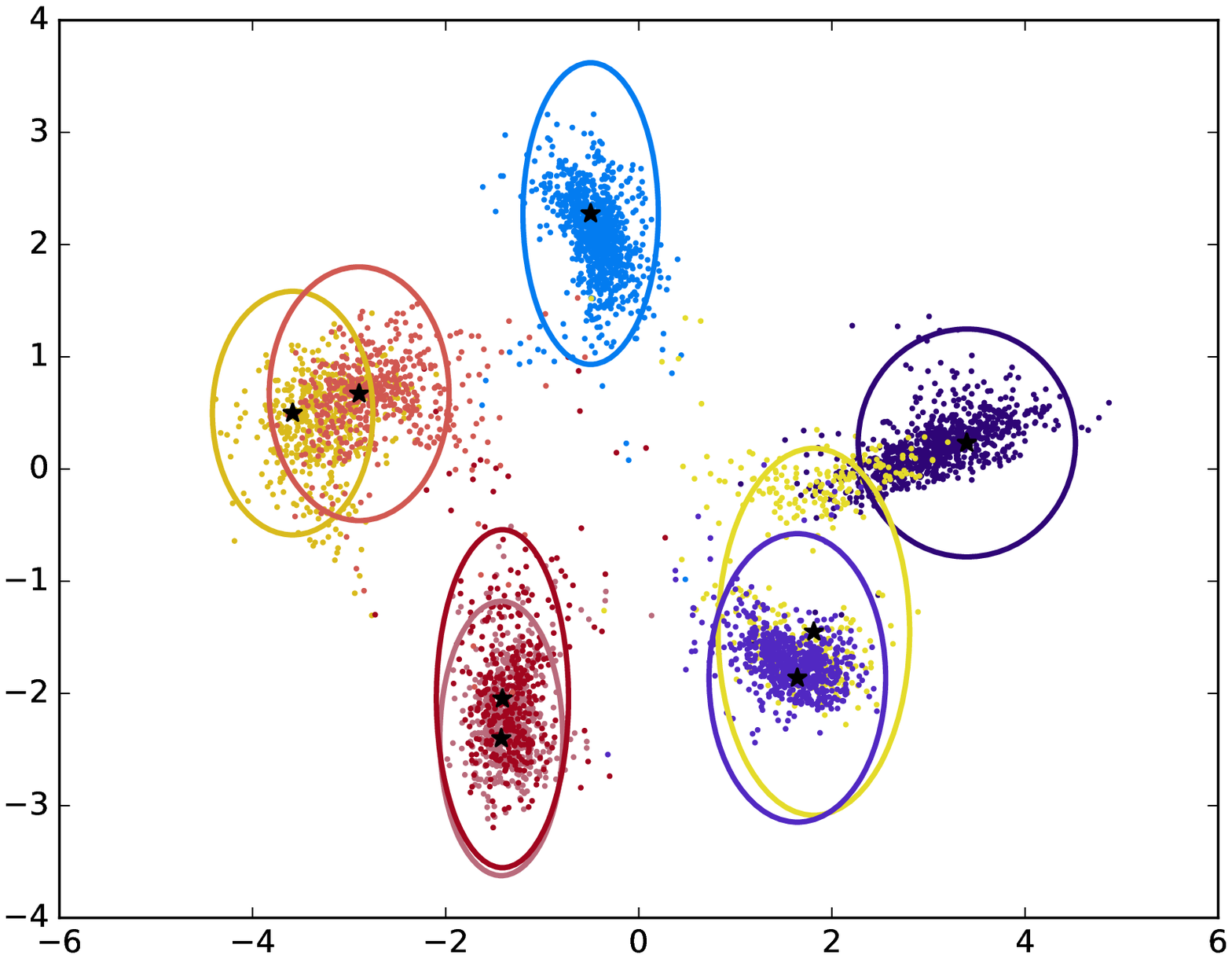}
	  \caption{Latent space, clusters spreading}
	  \label{fig:spread}
	\end{subfigure}%
	\begin{subfigure}{.33\textwidth}
    \centering
    \includegraphics[width=1.0\linewidth]{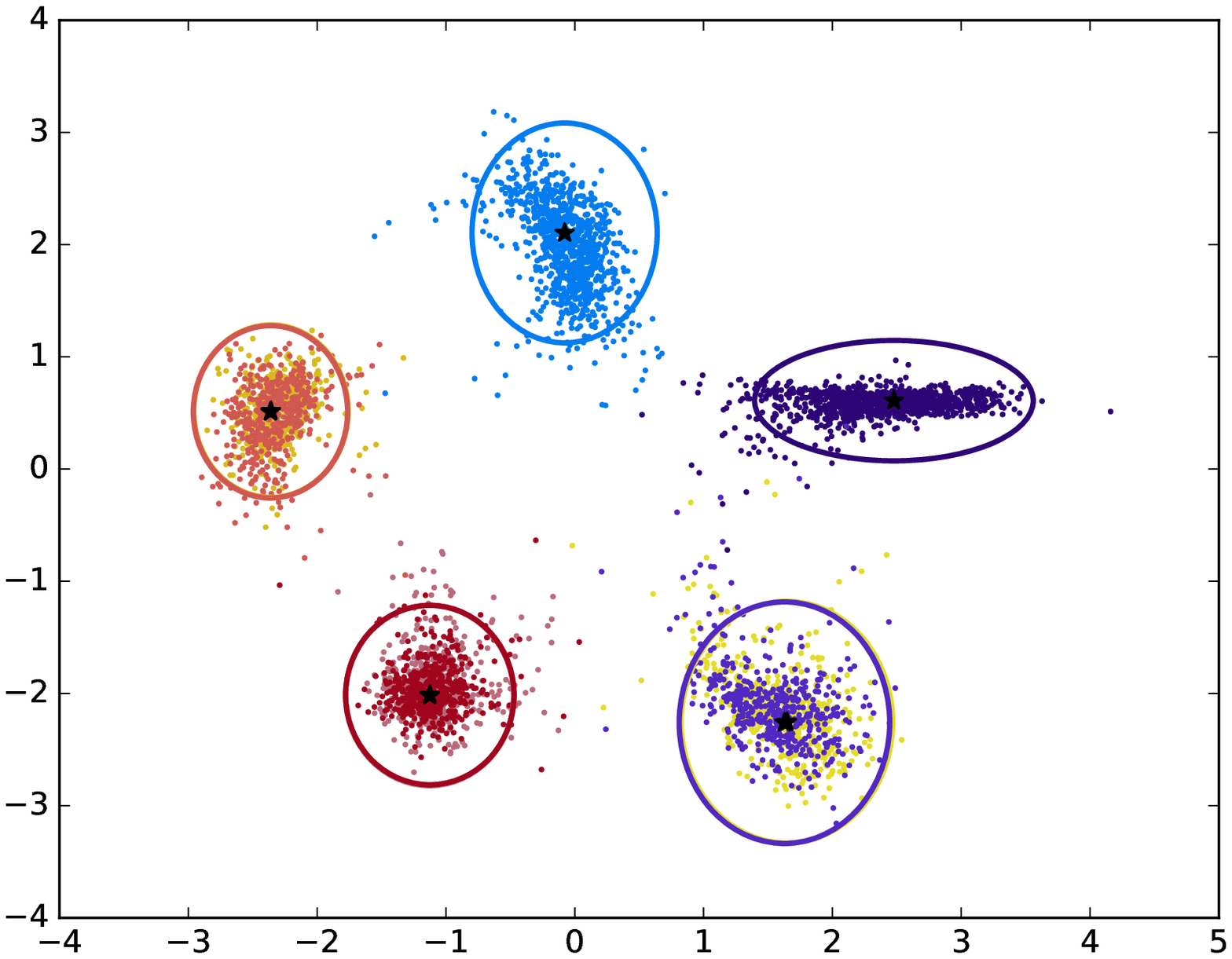}
	  \caption{Latent space, at convergence}
	  \label{fig:converge}
	\end{subfigure}

  \caption{\textbf{Visualisation of the synthetic dataset}: (a) Data is distributed with 5 modes on the 2 dimensional data space. (b) GMVAE learns the density model that can model data using a mixture of non-Gaussian distributions in the data space. (c) GMM cannot represent the data as well because of the restrictive Gaussian assumption. (d) GMVAE, however, suffers from over-regularisation and can result in poor minima when looking at the latent space. (e) Using the modification to the ELBO \citep{kingma2016improving} allows the clusters to spread out. (f) As the model converges the $z$-prior term is activated and regularises the clusters in the final stage by merging excessive clusters.}
  \label{fig:spiral}
\end{figure}

\begin{figure}[h]
\centering
%\framebox[4.0in]{$\;$}
%\fbox{\rule[-.5cm]{0cm}{4cm} \rule[-.5cm]{4cm}{0cm}}
  \begin{subfigure}{.5\textwidth}
		\centering
		\includegraphics[width=.9\linewidth]{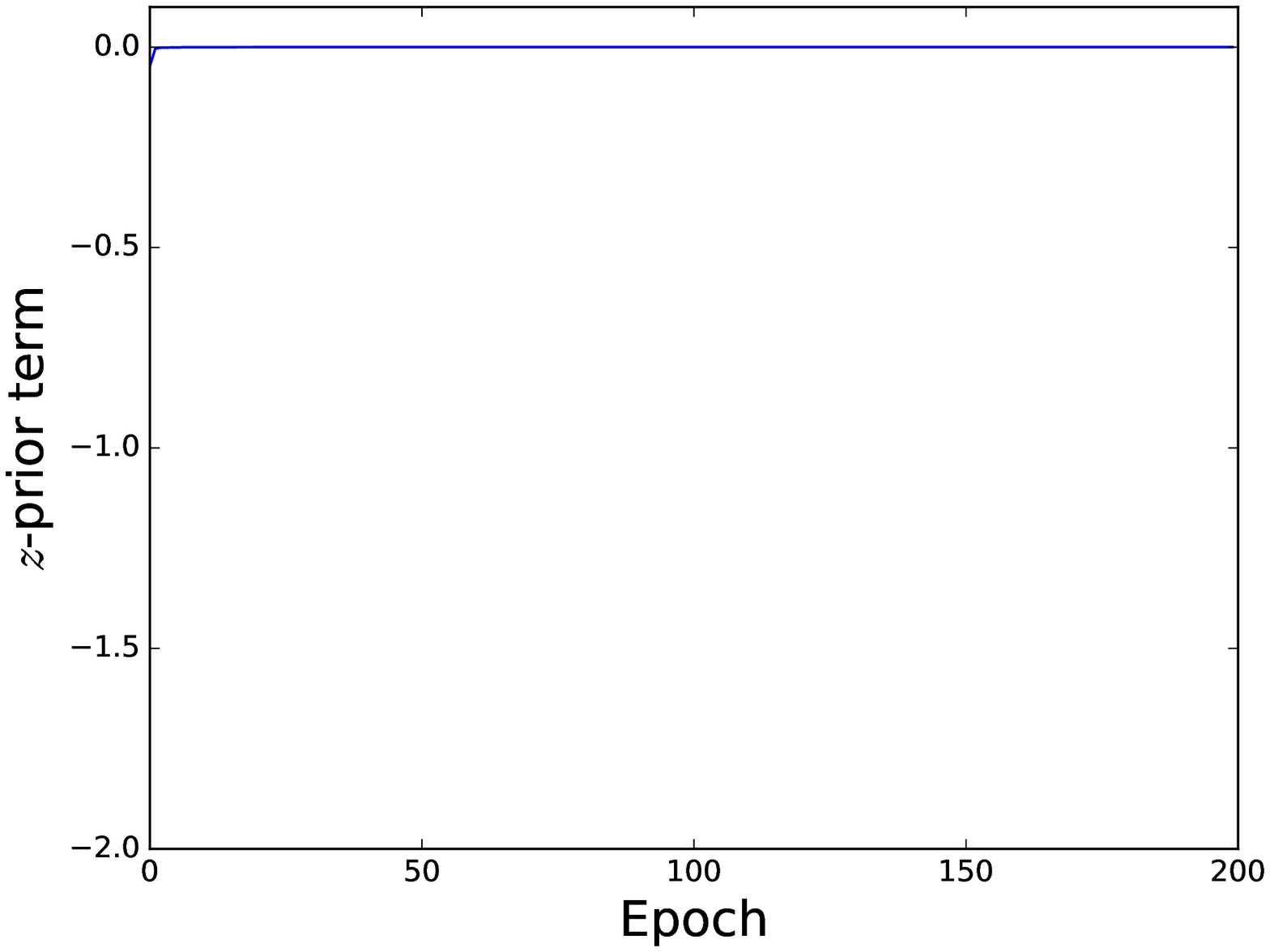}
		\caption{$z$-prior term with normal ELBO}
        \label{fig:zpriorfail}
	\end{subfigure}%
	\begin{subfigure}{.5\textwidth}
		\centering
		\includegraphics[width=.9\linewidth]{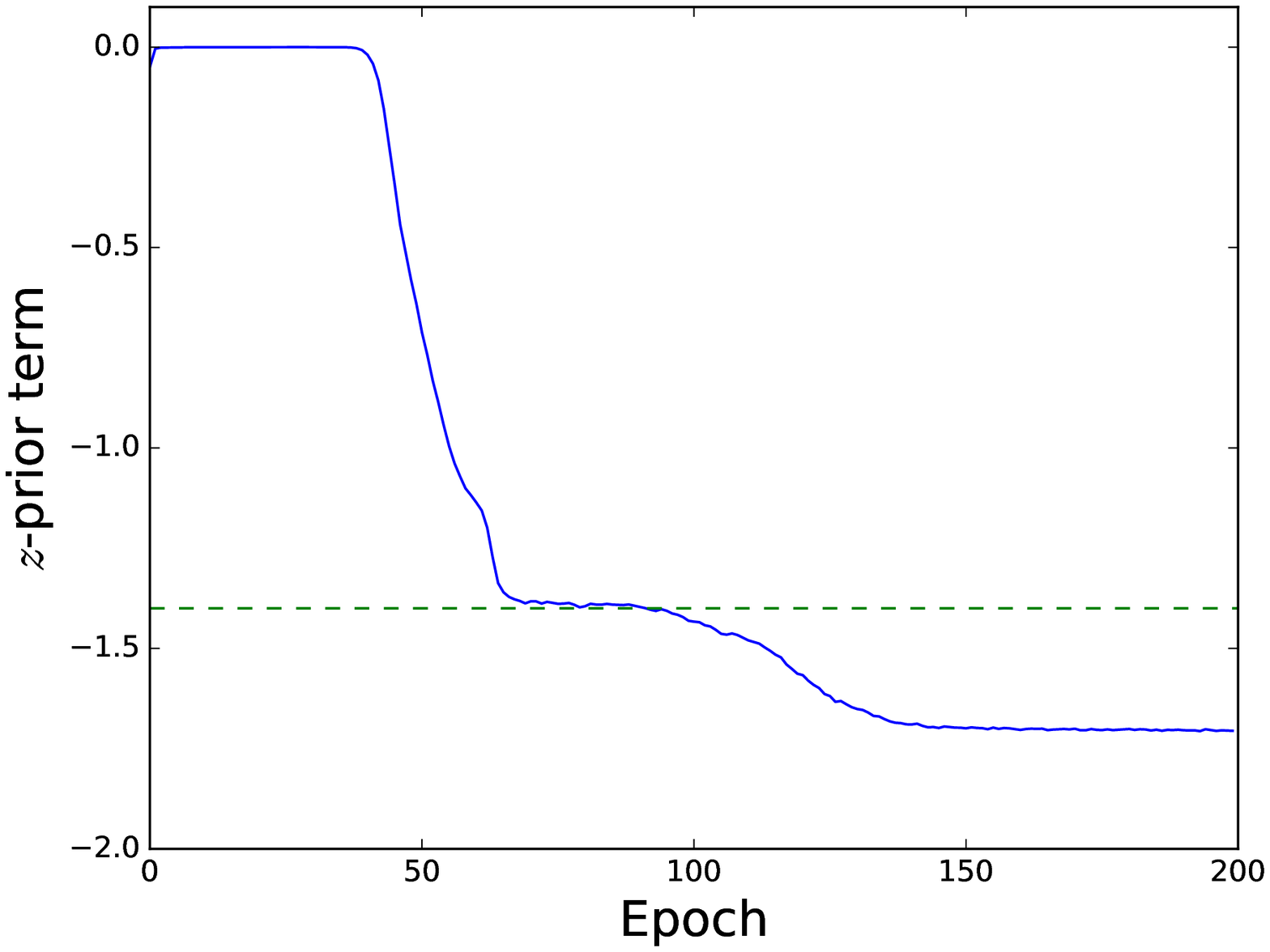}
		\caption{$z$-prior term with the modification}
        \label{fig:zprior}
	\end{subfigure}

\caption{\textbf{Plot of $z$-prior term}: (a) Without information constraint, GMVAE suffers from over-regularisation as it converges to a poor optimum that merges all clusters together to avoid the KL cost. (b) Before reaching the threshold value (dotted line), the gradient from the $z$-prior term can be turned off to avoid the clusters from being pulled together (see text for details). By the time the threshold value is reached, the clusters are sufficiently separated. At this point the activated gradient from the $z$-prior term only merges very overlapping clusters together. Even after activating its gradient the value of the $z$-prior continues to decrease as it is over-powered by other terms that lead to meaningful clusters and better optimum.}
\label{fig:zplot}
\end{figure}

Empirically, however, we have found this not to be the case. The latent representations that our model converges to merges all classes into the same large cluster instead of representing information about the different clusters, as can be seen in Figs.~\ref{fig:poorOptimal} and ~\ref{fig:zpriorfail}. As a result, each data point is equally likely to belong to any of clusters, rendering our latent representations completely uninformative with respect to the class structure.

We argue that this phenomenon can be interpreted as the result of over-regularisation by the $z$-prior term. Given that this quantity is driven up by the optimisation of KL term in the lower bound, it reaches its maximum possible value of zero, as opposed to decreasing with training to ensure encoding of information about the classes. We suspect that the prior has too strong of an influence in the initial training phase and drives the model parameters into a poor local optimum that is hard to be driven out off by the reconstruction term later on.

This observation is conceptually very similar to the over-regularisation problem encountered in regular VAEs and we thus hypothesize that applying similar heuristics should help alleviate the problem. We show in Fig.~\ref{fig:converge} that by using the previously mentioned modification to the lower-bound proposed by ~\cite{kingma2016improving}, we can avoid the over-regularisation caused by the $z$-prior. This is achieved by maintaining the cost from the $z$-prior at a constant value $\lambda$ until it exceeds that threshold. Formally, the modified $z$-prior term is written as:

\begin{equation}
\begin{aligned}
 \mathcal{L}_{z}' &= - \max (\lambda,  \mathbb{E}_{q(\pmb{x}|\pmb{y})q(\pmb{w}|\pmb{y})}\big[ KL(p_{\beta}(\pmb{z}| \pmb{x},\pmb{w})|| p(\pmb{z})) \big])
  \label{eq:z-prior_after}%
\end{aligned}
\end{equation}

This modification suppresses the initial effect of the $z$-prior to merge all clusters thus allowing them to spread out until the cost from the $z$-prior cost is high enough. At that point its effect is significantly reduced and is mostly limited to merging individual clusters that are overlapping sufficiently. This can be seen clearly in Figs.~\ref{fig:spread} and~\ref{fig:converge}. The former shows the clusters before the $z$-prior cost is taken into consideration, and as such the clusters have been able to spread out. Once the $z$-prior is activated, clusters that are very close together will be merged as seen in Fig.~\ref{fig:converge}.

Finally, in order to illustrate the benefits of using neural networks for the transformation of the distributions, we compare the density observed by our model (Fig.~\ref{fig:gmmDensity}) with a regular GMM (Fig.~\ref{fig:gmmDensity}) in data space. As illustrated by the figures, the GMVAE allows for a much richer, and thus more accurate representations than regular GMMs, and is therefore more successful at modelling non-Gaussian data. 

\subsection{Unsupervised Image Clustering}

We now assess the model's ability to represent discrete information present in the data on an image clustering task. We train a GMVAE on the MNIST training dataset and evaluate its clustering performance on the test dataset. To compare the cluster assignments given by the GMVAE with the true image labels we follow the evaluation protocol of \cite{makhzani2015adversarial}, which we summarise here for clarity. In this method, we find the element of the test set with the highest probability of belonging to cluster $i$ and assign that label to all other test samples belonging to $i$. This is then repeated for all clusters $i = 1, ..., K$, and the assigned labels are compared with the true labels to obtain an unsupervised classification error rate.

While we observe the cluster degeneracy problem when training the GMVAE on the synthetic dataset, the problem does not arise with the MNIST dataset. We thus optimise the GMVAE using the ELBO directly, without the need for any modifications. A summary of the results obtained on the MNIST benchmark with the GMVAE as well as other recent methods is shown in Table~\ref{table:mnist-cluster-table}. We achieve classification scores that are competitive with the state-of-the-art techniques\footnote{It is worth noting that shortly after our initial submission, Rui Shu published a blog post (http://ruishu.io/2016/12/25/gmvae/) with an analysis on Gaussian mixture VAEs. In addition to providing insightful comparisons to the aforementioned M2 algorithm, he implements a version that achieves competitive clustering scores using a comparably simple network architecture. Crucially, he shows that model M2 does not use discrete latent variables when trained without labels. The reason this problem is not as severe in the GMVAE might possibly be the more restrictive assumptions in the generative process, which helps the optimisation, as argued in his blog.}, except for adversarial autoencoders (AAE). We suspect the reason for this is, again, related to the KL terms in the VAE's objective. As indicated by Hoffman et al., the key difference in the adversarial autoencoders objective is the replacement of the KL term in the ELBO by an adversarial loss that allows the latent space to be manipulated more carefully~\citep{Hoffman2016ELBO}. Details of the network architecture used in these experiments can be found in Appendix \ref{ap:params}.

\begin{figure}[h]
\centering
 \includegraphics[trim={0.0cm 2.5cm 0.0cm 2.5cm},clip,width=.9\linewidth]{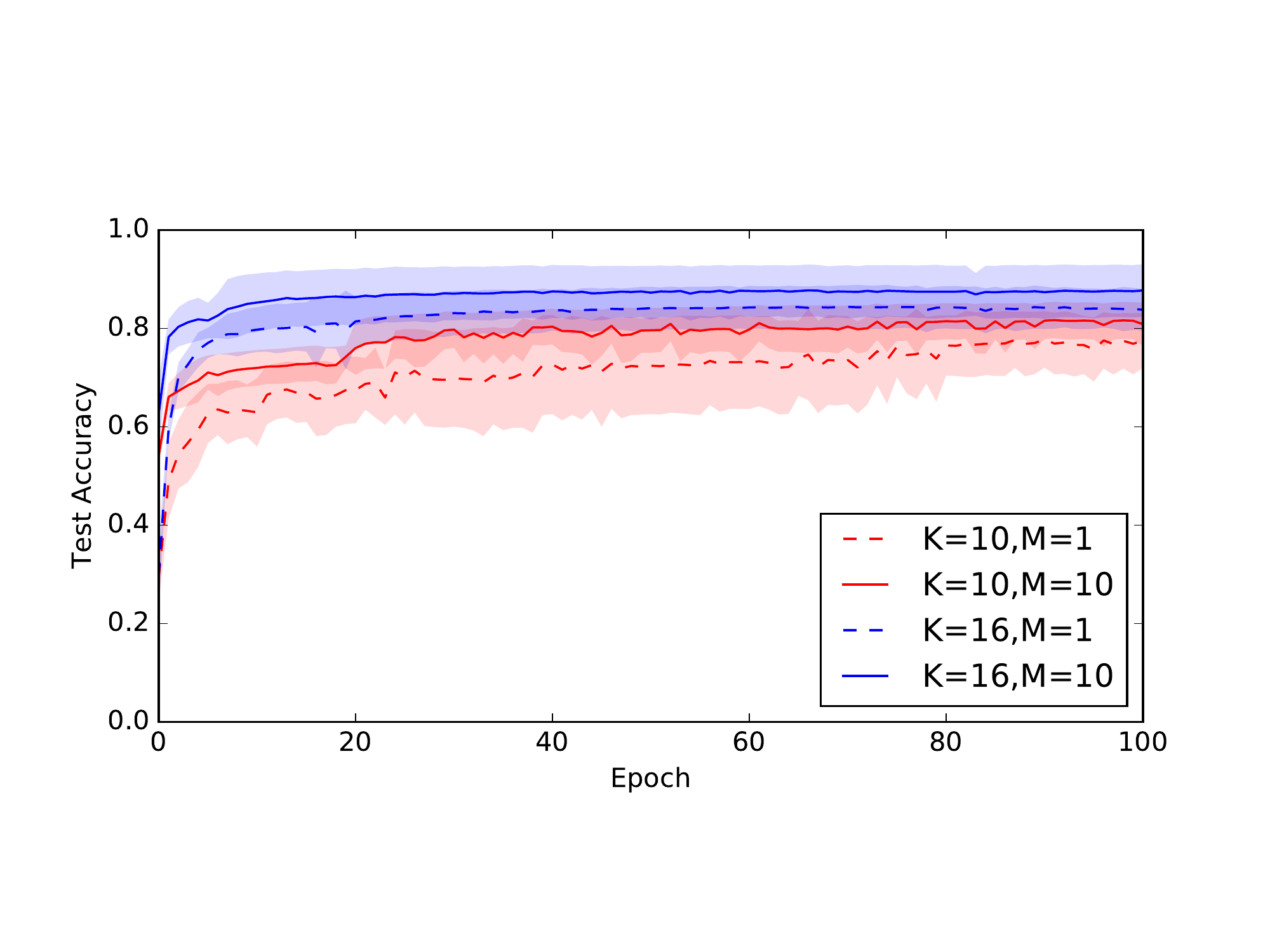}
\caption{\textbf{Clustering Accuracy with different numbers of clusters (K) and Monte Carlo samples (M)} : After only few epochs, the GMVAE converges to a solution. Increasing the number of clusters improves the quality of the solution considerably.}
\label{fig:acc}
\end{figure}

Empirically, we observe that increasing the number of Monte Carlo samples and the number of clusters makes the GMVAE more robust to initialisation and more stable as shown in Fig.~\ref{fig:acc}. If fewer samples or clusters are used then the GMVAE can occasionally converge faster to poor local minima, missing some of the modes of the data distribution.

\begin{table}[t]
\caption{Unsupervised classification accuracy for MNIST with different numbers of clusters (K) (reported as percentage of correct labels)}
\label{table:mnist-cluster-table}
\centering
\begin{tabular}{llll}
\multicolumn{1}{c}{\bf Method}  &\multicolumn{1}{c}{\bf K}		&\multicolumn{1}{c}{\bf Best Run } &\multicolumn{1}{c}{\bf Average Run}
\\ \hline \\
CatGAN \citep{springenberg2015unsupervised} &20	& 90.30 & - \\
AAE \citep{makhzani2015adversarial}&16& - & 90.45 $\pm$ 2.05\\
AAE \citep{makhzani2015adversarial} &30	&-& 95.90 $\pm$ 1.13 \\
DEC  \citep{xie2015unsupervised} &10 & 84.30 & - \\
\\
GMVAE (M = 1) & 10  & 87.31 &  77.78 $\pm$ 5.75 \\
GMVAE (M = 10) & 10 & 88.54  & 82.31 $\pm$ 3.75 \\
GMVAE (M = 1) &16 &	89.01 & 85.09 $\pm$ 1.99  \\
GMVAE (M = 10) &16 & 96.92  & 87.82 $\pm$ 5.33 	\\
GMVAE (M = 1) &30 &	95.84 & 92.77 $\pm$ 1.60 \\
GMVAE (M = 10) &30 & 93.22 & 89.27 $\pm$ 2.50	\\
\end{tabular}
\end{table}

\subsubsection{Image Generation}
So far we have argued that the GMVAE picks up natural clusters in the dataset, and that these clusters share some structure with the actual classes of the images. Now we train the GMVAE with $K = 10$ on MNIST to show that the learnt components in the distribution of the latent space actually represent meaningful properties of the data. First, we note that there are two sources of stochasticity in play when sampling from the GMVAE, namely

\begin{enumerate}
\item Sampling $\pmb{w}$ from its prior, which will generate the means and variances of $\pmb{x}$ through a neural network $\beta$; and
\item Sampling $\pmb{x}$ from the Gaussian mixture determined by $\pmb{w}$ and $\pmb{z}$, which will generate the image through a neural network $\theta$.
\end{enumerate}

\begin{figure}[h]
\centering
	\begin{subfigure}{.5\textwidth}
		\centering
		\includegraphics[width=.9\linewidth]{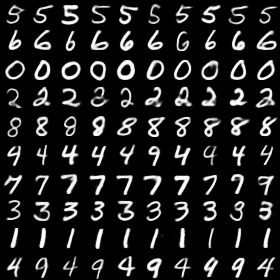}
		\caption{Varying $z$}
        \label{fig:mnist_gen}
	\end{subfigure}%
	\begin{subfigure}{.5\textwidth}
		\centering
		\includegraphics[width=.9\linewidth]{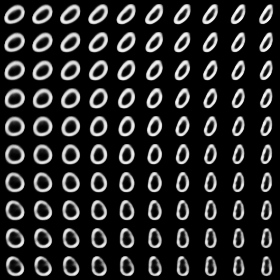}
		\caption{Varying $w$}
        \label{fig:mnist_style}
	\end{subfigure}
\caption{\textbf{Generated MNIST samples}: (a) Each row contains 10 randomly generated samples from different Gaussian components of the Gaussian mixture. The GMVAE learns a meaningful generative model where the discrete latent variables $z$ correspond directly to the digit values in an unsupervised manner. (b) Samples generated by traversing around $w$ space, each position of $w$ correspond to a specific style of the digit.}
\label{fig:mnist}
\end{figure}

In Fig.~\ref{fig:mnist_gen} we explore the latter option by setting $\pmb{w} = 0$ and sampling multiple times from the resulting Gaussian mixture. Each row in Fig.~\ref{fig:mnist_gen} corresponds to samples from a different component of the Gaussian mixture, and it can be clearly seen that samples from the same component consistently result in images from the same class of digit. This confirms that the learned latent representation contains well differentiated clusters, and exactly one per digit. Additionally, in Fig.~\ref{fig:mnist_style} we explore the sensitivity of the generated image to the Gaussian mixture components by smoothly varying $\pmb{w}$ and sampling from the same component. We see that while $\pmb{z}$ reliably controls the class of the generated image, $\pmb{w}$ sets the ``style'' of the digit.

\begin{figure}[h]
\centering
%\framebox[4.0in]{$\;$}
%\fbox{\rule[-.5cm]{0cm}{4cm} \rule[-.5cm]{4cm}{0cm}}
\includegraphics[width=0.5\textwidth]{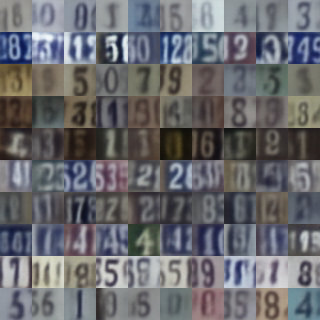}
 \caption{\textbf{Generated SVHN samples}: Each row corresponds to 10 samples generated randomly from different Gaussian components. GMVAE groups together images that are visually similar.}
\label{fig:SVHN}
\end{figure}

Finally, in Fig.~\ref{fig:SVHN} we show images sampled from a GMVAE trained on SVHN, showing that the GMVAE clusters visually similar images together.

\section{Conclusion}

We have introduced a class of variational autoencoders in which one level of the latent encoding space has the form of a Gaussian mixture model, and specified a generative process that allows us to formulate a variational Bayes optimisation objective. We then discuss the problem of over-regularisation in VAEs. In the context of our model, we show that this problem manifests itself in the form of cluster degeneracy. Crucially, we show that this specific manifestation of the problem can be solved with standard heuristics. 

We evaluate our model on unsupervised clustering tasks using popular datasets and achieving competitive results compared to the current state of the art. Finally, we show via sampling from the generative model that the learned clusters in the latent representation correspond to meaningful features of the visible data. Images generated from the same cluster in latent space share relevant high-level features (e.g. correspond to the same MNIST digit) while being trained in an entirely unsupervised manner.

It is worth noting that GMVAEs can be stacked by allowing the prior on $w$ to be a Gaussian mixture distribution as well. A deep GMVAE could scale much better with number of clusters given that it would be combinatorial with regards to both number of layers and number of clusters per layer. As such, while future research on deep GMVAEs for hierarchical clustering is a possibility, it is crucial to also address the enduring optimisation challenges associated with VAEs in order to do so.

\subsubsection*{Acknowledgments}
We would like to acknowledge the NVIDIA Corporation for the donation of a GeForce GTX Titan Z used in our experiments. We would like to thank Jason Rolfe, Rui Shu and the reviewers for useful comments. Importantly, we would also like to acknowledge that the variational family which we used throughout this version of the paper was suggested by an anonymous reviewer. 

\bibliography{references}
\bibliographystyle{iclr2017_conference}

%%%%%%%%%%%%%%%%%%%%%%%%%%%%%%%%%%%%%%%%%%%%%%%%
%%%%%%%%%%%%%%%%%%%%%%%%%%%%%%%%%%%%%%%%%%%%%%%%
\appendix
\counterwithin{table}{section}
\section{Network Parameters}
\label{ap:params}
For optimisation, we use Adam \citep{kingma2014adam} with a learning rate of \num{e-4} and standard hyperparameter values $\beta_1 = 0.9$, $\beta_2 = 0.999$ and $\epsilon = 10^{-8}$. The model architectures used in our experiments are shown in Tables \ref{table:architecture1}, \ref{table:architecture2} and \ref{table:architecture3}.

\begin{table}[ht]
\caption{\textbf{Neural network architecture models of $q_{\phi}(\pmb{x},\pmb{w})$}: The hidden layers are shared between $q(\pmb{x})$ and $q(\pmb{w})$, except the output layer where the neural network is split into 4 output streams, 2 with dimension $N_x$ and the other 2 with dimension $N_w$. We exponentiate the variance components to keep their value positive. An asterisk (*) indicates the use of batch normalization and a ReLU nonlinearity. For convolutional layers, the numbers in parentheses indicate stride-padding.}
\label{table:architecture1}
\centering
\begin{tabular}{llll}
\multicolumn{1}{c}{\bf Dataset} &\multicolumn{1}{c}{\bf Input}  &\multicolumn{1}{c}{\bf Hidden}		&\multicolumn{1}{c}{\bf Output }
\\ \hline \\
Synthetic  & 2  &  fc 120 ReLU 120 ReLU  				            &  $N_w$ = 2, $N_w$ = 2 (Exp), \\
  & & 																						&	 $N_x$ = 2, $N_x$ = 2 (Exp) \\
\\ \hline \\
MNIST  &28x28 & conv 16x6x6* (1-0) 32x6x6* (1-0)  & $N_w$ = 150, $N_w$ = 150 (Exp),\\
  & &64x4x4* (2-1) 500* 														&	$N_x$ = 200, $N_x$ = 200 (Exp)\\
\\	\hline \\
SVHN  &32x32 & conv 64x4x4* (2-1) 128x4x4* (2-1)  & $N_w$ = 150, $N_w$ = 150 (Exp),\\
  & & 	246x4x4* (2-1) 500*											 	&	$N_x$ = 200, $N_x$ = 200 (Exp) \\
\\ \hline \\
\end{tabular}
\end{table}

\begin{table}[ht]
\caption{\textbf{Neural network architecture models of $p_{\beta}(\pmb{x}| \pmb{w}, \pmb{z})$}: The output layers are split into $2K$ streams of output, where $K$ streams return mean values and the other $K$ streams output variances of all the clusters.}
\label{table:architecture2}
\centering
\begin{tabular}{llll}
\multicolumn{1}{c}{\bf Dataset} &\multicolumn{1}{c}{\bf Input}  &\multicolumn{1}{c}{\bf Hidden}		&\multicolumn{1}{c}{\bf Output }
\\ \hline \\
Synthetic  & 2  &  fc 120 Tanh  &   $\{ N_x = 2\}_{2K}$  \\
\\ \hline \\
MNIST  &150 & fc 500 Tanh &   $\{ N_x = 200\}_{2K}$\\
\\	\hline \\
SVHN  &150 & fc 500 Tanh  &   $\{ N_x = 200\}_{2K}$ \\

\\ \hline \\
\end{tabular}
\end{table}

\begin{table}[ht]
\caption{\textbf{Neural network architecture models of $p_{\theta}(\pmb{y}| \pmb{x})$}: The network outputs are Gaussian parameters for the synthetic dataset and Bernoulli parameters for MNIST and SVHN, where we use the logistic function to keep value of Bernoulli parameters between 0 and 1. An asterisk (*) indicates the use of batch normalization and a ReLU nonlinearity. For convolutional layers, the numbers in parentheses indicate stride-padding.}
\label{table:architecture3}
\centering
\begin{tabular}{llll}
\multicolumn{1}{c}{\bf Dataset} &\multicolumn{1}{c}{\bf Input}  &\multicolumn{1}{c}{\bf Hidden}		&\multicolumn{1}{c}{\bf Output }
\\ \hline \\
Synthetic  & 2  &  fc 120 ReLU 120 ReLU 					            & $\{2\}_2$  \\
\\ \hline \\
MNIST  & 200 & 500* full-conv 64x4x4* (2-1) 32x6x6* (1-0)  & 28x28 (Sigmoid) \\
  		 & 		 & 		16x6x6* (1-0) 								 &	  \\
			 & &  												  					 &    \\
\\	\hline \\
SVHN  & 200 & 500* full-conv 246x4x4* (2-1) 128x4x4* (2-1)  &  32x32 (Sigmoid) \\
  		& & 	64x4x4* (2-1) 								 	&	  \\
			& &  											 								 &   \\
\\ \hline \\
\end{tabular}
\end{table}

\end{document}